\pdfoutput=1

\documentclass[11pt]{article}

\usepackage{naacl2021}

\usepackage{times}
\usepackage{latexsym}

\usepackage[T1]{fontenc}

\usepackage[utf8]{inputenc}

\usepackage{microtype}

\usepackage{nccmath}
\usepackage{amssymb}
\usepackage{bbold}
\usepackage{todonotes}
\usepackage{microtype}
\usepackage{comment}

\newcommand{\rc}{{\sc MRC}}
\newcommand{\semanticParsing}{IS-SP}
\newcommand{\GAAMA}{MRC}
\newcommand{\GAAMAXXL}{MRC$_{\text{xxl}}$}
\newcommand{\RCIinteraction}{RCI$_{\text{inter}}$}
\newcommand{\RCIrepresentation}{RCI$_{\text{repr}}$}
\newcommand{\RCIXXL}{RCI$_{\text{xxl}}$}
\newcommand{\TaBERT}{{\sc TaBert}}
\newcommand{\TAPAS}{{\sc TaPas}}

\newcommand{\eg}{\textit{e.g.}}

\title{Capturing Row and Column Semantics in \\
Transformer Based Question Answering over Tables}

\author{
Michael Glass$^{1}$,
Mustafa Canim$^{1}$, 
Alfio Gliozzo$^{1}$, \\
\textbf{Saneem Chemmengath$^{1}$,
Vishwajeet Kumar$^{1}$,
Rishav Chakravarti$^{1}$, 
Avi Sil$^{1}$,
Feifei Pan$^{2}$,} \\
\textbf{Samarth Bharadwaj$^{1}$, 
Nicolas Rodolfo Fauceglia$^{1}$
} \\
$^{1}$ IBM Research AI 
$^{2}$ Rensselaer Polytechnic Institute\\
mrglass@us.ibm.com, mustafa@us.ibm.com, gliozzo@us.ibm.com, \\ saneem.cg@in.ibm.com, vishk024@in.ibm.com, rchakravarti@us.ibm.com, \\ avi@us.ibm.com, panf2@rpi.edu, samarth.b@in.ibm.com, nicolas.fauceglia@ibm.com\\
    }

\begin{document}
\maketitle
\begin{abstract}
Transformer based architectures are recently used for the task of answering questions over tables. In order to improve the accuracy on this task, specialized pre-training techniques have been developed and applied on millions of open-domain web tables.
In this paper, we propose two novel approaches demonstrating that one can achieve superior performance on table QA task without even using any of these specialized pre-training techniques. The first model, called \emph{RCI interaction}, leverages a transformer based architecture that independently classifies rows and columns to identify relevant cells. While this model yields extremely high accuracy at finding cell values on recent benchmarks, a second model we propose, called \emph{RCI representation}, provides a significant efficiency advantage for online QA systems over tables by materializing embeddings for existing tables. Experiments on recent benchmarks prove that the proposed methods can effectively locate cell values on tables (up to $\sim$98\% Hit@1 accuracy on WikiSQL lookup questions). Also, the interaction model outperforms the state-of-the-art transformer based approaches, pre-trained on very large table corpora (\TAPAS{} and \TaBERT{}), achieving $\sim$3.4\% and $\sim$18.86\% additional precision improvement on the standard WikiSQL benchmark\footnote{The source code and the models we built are available at https://github.com/IBM/row-column-intersection.}.

\end{abstract}

\section{Introduction}
\label{sec:intro}
Tabular data format is a commonly used layout in domain specific enterprise documents as well as open domain webpages to store structured information in a compact form~\cite{pasupat2015compositional, canim2019}. In order to make use of these resources, many techniques have been proposed for the retrieval of tables~\cite{cafarella2008webtables,ZhangAdHoc,venetis2011recovering,www2020,sun2016table}. Given a large corpus of documents, the goal in these studies is to retrieve top-k relevant tables based on given keyword(s). The user is then expected to skim through these tables and locate the relevant cell values which is a tedious and time consuming task. More recently, popular search engines 
made significant improvement in understanding natural language questions and finding the answers within passages, owing to the developments in transformer based machine reading comprehension (MRC) systems~\cite{Rajpurkar_2016,rajpurkar2018know, Kwiatkowski2019NaturalQA, pan2019frustratingly, albert-synth-data}.
One natural extension of these systems is to answer questions over tables. 
These questions are broadly classified into two types: \emph{Lookup} and \emph{Aggregation}. \emph{Lookup} questions require returning exact strings from tables such as cell values whereas \emph{Aggregation} questions are executed by performing an arithmetic operation on a subset of the column cells, such as \emph{Min(), Max(), Average() and Count()}. For look-up questions, the users can verify if the returned cell values from the table(s) are correct, while this is not applicable for \emph{Aggregation} questions because a scalar value is returned as an answer. Our primary focus in this paper is on \emph{Lookup} questions since the answers are verifiable by users although our proposed techniques outperform the state-of-the-art (SOTA) approaches on both question types.

\begin{figure}[thb]
   \centering
   \includegraphics[width=1\linewidth]{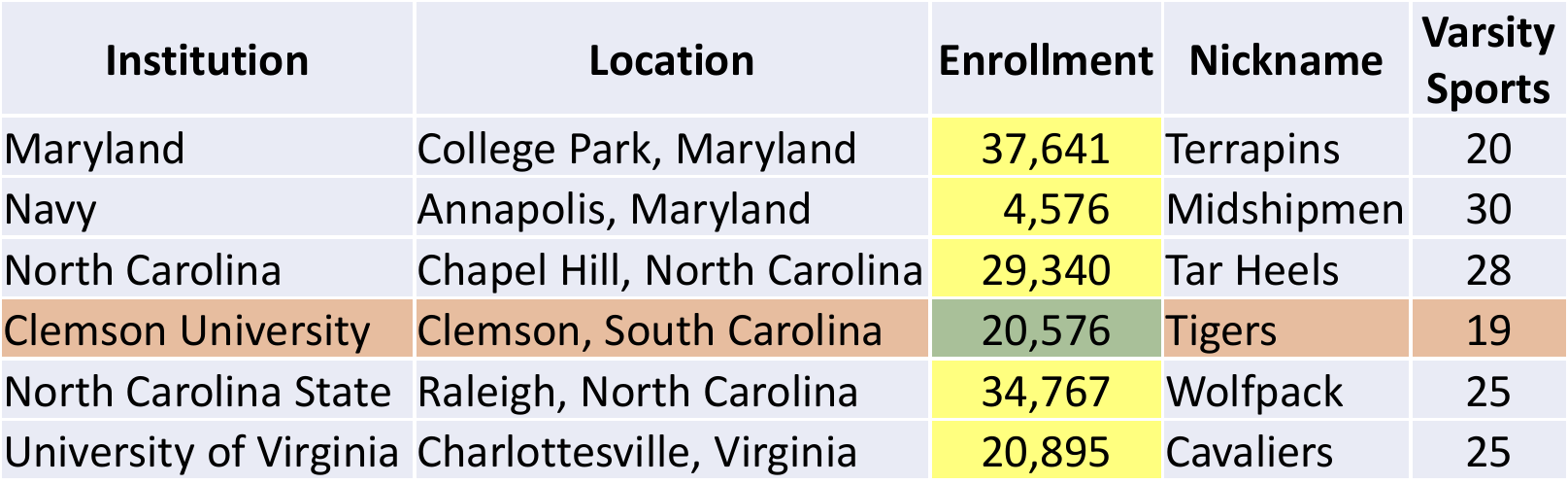}
   \caption{Answering a question ``What is the Clemson Tiger's enrollment?'' over a table with a heatmap}
   \label{fig.heatmap}
\end{figure}

In this paper, we propose a new approach to table QA that independently predicts the probability of containing the answer to a question in each row and column of a table.
By taking the \textbf{R}ow and \textbf{C}olumn \textbf{I}ntersection (RCI) of these probabilistic predictions, RCI gives a probability for each cell of the table.
These probabilities are either used to answer questions directly or highlight the relevant regions of tables as a heatmap, helping users to easily locate the answers over tables (See Figure~\ref{fig.heatmap} for a question answered with the help of a heatmap).
We developed two models for RCI, called \emph{RCI interaction} and \emph{RCI representation}.

In order to evaluate these approaches, we also propose a weakly supervised MRC system as a strong baseline to identify / "read" relevant cells of a table. In this baseline approach, we convert tables into passages and extract a relevant span of text within these passages. 

The interaction model is designed to provide very high accuracy on finding cell values over tables for a given natural language question. 
We demonstrate that without even using any specialized pre-trained models, we can achieve up-to $\sim$98\% Hit@1 accuracy on finding cell values of tables for lookup questions from the WikiSQL benchmark. 
Also, the interaction model outperforms the state-of-the-art transformer based approaches, \TAPAS~\cite{tapas} and \TaBERT~\cite{tabert}, achieving $\sim$3.4\% and $\sim$18.86\% additional precision improvement on the standard WikiSQL benchmark, containing both lookup and aggregation questions.

While the interaction model yields very high accuracy on the benchmarks, the representation model has the advantage of pre-computing the embeddings for all tables in a corpus and storing them for online query processing. Once a user query is received, the most relevant tables can be retrieved from a table retrieval system and relevant cell values can be highlighted using the existing embeddings of the tables, resulting in less computation per received user query, as opposed to running tables over expensive transformer architecture for every received query. 

The specific contributions of this paper are as follows:
\begin{itemize}
  \item \textbf{An MRC based strong baseline for table QA task:} We investigate a transfer learning approach by utilizing a fully supervised reading comprehension system built on top of a large pre-trained language model. Specifically, it is first fine-tuned on SQuAD then on Natural Questions and lastly trained on the table datasets. The final model is used to identify relevant cells of a table for a given natural language question. 

  \item \textbf{A transformer based interaction model for the table QA task:} We propose a model for table QA task that concatenates a textual representation of each row (or column) to the text of the question and classifies the sequence pair as positive (the row/column contains the answer) or negative (the row/column does not contain the answer). The proposed approach yields very high accuracy on our benchmarks, outperforming the SOTA models.
  
  \item \textbf{A transformer based representation model for the table QA task:} We propose a representation model that builds vector representations of the question and each row (or column) to compare the resulting vectors to determine if the row (or column) contains the answer. The proposed approach is preferred for efficiency purposes on online table retrieval systems since it enables materializing embeddings for existing tables and re-using them during online question answering over multiple tables.
\end{itemize}

In the following sections, we first review the prior work on QA systems over tables as well as table search from large corpora in Section~\ref{sec:relatedwork}.  We then describe a weakly supervised machine reading comprehension (MRC) system as a baseline that is capable of answering questions over tables in Section~\ref{sec:MRC}. In Section~\ref{sec:model_rci}, we introduce two models that decompose TableQA as the intersection between rows and columns of a table using a transformer architecture. Experimental results are reported and discussed in Section~\ref{sec:eval} and finally Section~\ref{sec:conclusion} concludes the paper and discusses the future work.
\section{Related Work}\label{sec:relatedwork}

\textbf{QA from text:} There is plenty of work on QA from plain text \cite{brill2002analysis,lin2007exploration,pacsca2003open,Kwiatkowski2019NaturalQA,pan2019frustratingly}. Typical strategies rely on token overlap between the question and passage text either based on a bag of word statistics or contextualized language model representations. In either case, tabular structure is not leveraged to capture semantic relationships between rows and columns. As we show in Section~\ref{sec:eval}, these strategies are insufficient for answering questions over tables with high precision.

\noindent\textbf{QA over tables:} Our work mostly relates to the previous research on QA over tables \cite{pasupat2015compositional,sun2016table, dasigi2019semparse}. They center around answering factoid questions and return the exact cell of a table that answers the query. We briefly describe here how these works are different. \citet{pasupat2015compositional} assume access to the `gold' table that contains the answer to the input question. They build a semantic parser that parses the query to a logical form. They likewise convert the table into a knowledge-graph and execute the logical form on it to get the answer. A more advanced semantic parsing based methodology has been recently proposed by~\citet{dasigi2019semparse}. This system is pre-trained on WikiTablesQuestions~\cite{pasupat2015compositional}. The proposed approach leverages an LSTM encoder-decoder model where tables are first converted to a knowledge-graph and word tokens in the questions are linked to table entities (columns and cells). The questions and linked table entities are then encoded into representation vectors which are decoded to executable $\lambda$-DCS logical forms. This logical forms are executed over a knowledge graph to get answer predictions. Our approach is different, since we do not convert natural language questions into logical forms and execute them on tables. Instead, we leverage transformer architectures pre-trained on large corpora and further trained on finding cell values on tables. In Section~\ref{sec:eval}, we show that we achieve significant improvement over this approach without using any semantic parser technique. 

\citet{sun2016table} focus on the table retrieval problem over table corpora by leveraging the content of cell values and headers. For a given query, they extract answers from millions of tables in the provided corpus. They construct a unified chain representation of both the input question and the table cells and then find the table cell chain that best matches the question chain. As opposed to this work, we primarily focus on answering questions over a single table rather than the retrieval of top-k tables from a corpus.

More recently, transformer based pre-training approaches have been introduced in \TaBERT{} \cite{tabert} and \TAPAS{} \cite{tapas} to improve accuracy for table QA. \TaBERT{} has been pre-trained on 26 million tables and NL sentences extracted from Wikipedia and WDC WebTable Corpus~\cite{tabert}. The model can be plugged into a neural semantic parser as an encoder to provide contextual embeddings for tables. \citeauthor{tapas} on the other hand, claim that semantic parsers incur an extra overhead of computing intermediate logical representations which can be avoided by leveraging fine-tuned models to answer questions over tables. The model in \TAPAS{} has been pre-trained on about 6 million tables extracted from Wikipedia content. Our work is different from both \TAPAS{} and \TaBERT{}. First and foremost, our focus in this paper is not on pre-training a new model for table QA, but rather on leveraging the existing language models to find the connection between a question and table columns/rows with very high accuracy. 
Second, our goal is to provide a heatmap over tables on an end-to-end table retrieval system to help users to quickly identify the regions of tables where the answers would most likely appear. Because the transformer architectures are quite expensive to query, the representation model we propose radically reduces the computational overhead during online query processing.

\noindent \textbf{Table search over the web:} Another active research area in NLP is searching over web tables. There are numerous search algorithms that have been explored such as keyword search~\cite{cafarella2008webtables,ZhangAdHoc,venetis2011recovering, www2020}, retrieve similar tables~\cite{das2012finding}, retrieve tables based on column names~\cite{pimplikar2012answering} and adding new columns to existing entity lists~\cite{yakout2012infogather,zhang2013infogather+}. This thread of work focuses on retrieval of top-k tables with high precision from large corpora, rather than finding relevant rows and columns within tables.  
\section{MRC Model}
\label{sec:MRC}
 
We provide a brief description of our underlying Machine Reading Comprehension (MRC) model architecture, which we use as a strong baseline. The architecture is inspired by \cite{alberti2019bert,pan2019frustratingly,glass2019span} and direct interested readers to their papers for more details. 
Our MRC model follows the approach introduced by \cite{Devlin2018BERTPO} of starting with a pre-trained transformer based language model (LM) and then fine-tuning MRC specific feed-forward layers on both general question answering datasets (SQuAD 2.0 and NQ) as well as the table specific question answers associated with the datasets in Section~\ref{sec:eval}. 

We use ALBERT \cite{lan2019albert} as the underlying LM similar to models which achieve SOTA on the SQuAD 2.0 leaderboard \cite{zhang2020retrospective, zhang2019sg} at the time of writing. 
More specifically, we show results starting from the weights and dimensions of the \textit{base v2} version (25M parameters) of the LM shared by \cite{Lan2020ALBERT}. 
We also experiment with the \textit{xxlarge v2} version (235M parameters) as well.
The input to the model is a token sequence ($\mathbf{X}$) consisting of a question, passage, and special markers (a $[CLS]$ token for answerability classification and $[SEP]$ tokens to dileneate between the query and passage). The input token sequence is passed through a deep Transformer \cite{Vaswani_2017} network to output a sequence of contextualized token representations $\mathbf{H}$.

\rc{} then adds two dense layers followed by a \emph{softmax}:

\vspace{-6 mm}
\begin{align*}
    \boldsymbol{\alpha}_b = softmax(\mathbf{W}_1 \mathbf{H}),\\
    \boldsymbol{\alpha}_e = softmax(\mathbf{W}_2 \mathbf{H}),
\end{align*}

where $\mathbf{W}_1$, $\mathbf{W}_2 \in \mathbb{R}^{1\times D_e}$. $D_e$ denotes the dimensionality of the embeddings 
(
768 for \textit{base v2} ). $\boldsymbol{\alpha}_b^t$ and $\boldsymbol{\alpha}_e^t$ denote the probability of the $t^{th}$ token in the sequence being the answer beginning and end, respectively.  

The model is trained using binary cross-entropy loss at each token position based on whether or not the annotated correct answer begins or ends at the  $t^{th}$ token. Unanswerable questions have their begin and end offsets set to the $[CLS]$ token position.

At prediction time, a score is calculated for each possible span by summing the $\boldsymbol{\alpha}_b^{t_j}$ and $\boldsymbol{\alpha}_e^{t_i}$ at each possible $i$ and $j$ combination to identify the max scoring answer span.  The sum of the $\boldsymbol{\alpha}_b^{[CLS]}$ and $\boldsymbol{\alpha}_e^{[CLS]}$ is then subtracted from this max scoring answer span to produce a final score that can be used for thresholding (i.e., deciding whether to predict an answer or refrain from answering a question). A few modifications are made in line with \cite{alberti2019bert} to use \rc{} for the NQ dataset which introduces additional answer types $[short, long, yes, no, null]$. Refer to the appendix for these details.

We fine-tune the model with the SQuAD 2.0 dataset and then the NQ dataset in line with \cite{pan2019frustratingly,sspt}, to produce a generic RC model comparable to the current SOTA. We then train for an additional epoch on the subset of NQ which consists of short answer questions that need to be answered by lookup inside an HTML table. This is about $5\%$  of the total NQ  data ($\sim15,500$ question-answer pairs). Note that in these cases, the input ``passage'' text consists of textual representation of tables (i.e., we introduce tabs between columns and new line characters between rows); so it is devoid of true row and column structure. 
This pre-training and task adaptation strategy is inline with prior art~\cite{gururangan2020dont} in adapting transformers.
Simpler pre-training strategies (e.g. relying only on SQuAD 2.0 or skipping the table specific epoch of training) were tried and found to provide similar, but generally worse, performance. So those are excluded from Section~\ref{sec:eval} for brevity.

Finally, we fine-tune (i.e., train for an additional epoch) on the training examples (table-question pairs) associated with the appropriate evaluation data sets described in Section~\ref{sec:eval}. During this step we do not have access to exact span offsets in the ground truth annotations and, instead, use weak supervision by matching the first occurrence of the answer text within the textual representation of the table\footnote{We provide the hyperparameters for the training process in the appendix.}.  
\section{RCI Model Architecture}
\label{sec:model_rci}
The Row-Column Intersection model (RCI) is motivated by the idea of decomposing lookup Table QA into two operations: the column selection and the row selection. Combining the predicted answer probability of each row and the probability of each column gives a score for all cells in the table.  The highest scoring cell may then be returned as an answer, or highlighting may be applied to the table to aid a user in locating the relevant information.  Unlike the pointer network of an adapted Machine Reading Comprehension system (described in Section~\ref{sec:MRC}), the RCI model always gives a ranked list of cells rather than answer spans that may cross cell boundaries.

We observe that the process of identifying the correct column is often about matching the column header and the type of values in the column to the expected answer type of the question.  For example in Table~\ref{tbl.example1}, the question has a lexical answer type of `party' and the column header for the correct column is `Party' and contains values that are political parties.

Identifying the correct row is often more difficult. In the example given in Table~\ref{tbl.example1}, it is sufficient to match either of the names in the question to the value in the `Name' column of the row.
Note that with weak supervision~\cite{min2019discrete} we do not know the correct row, so all occurrences of `Pro-Administration' are considered correct.

\begin{table}[th!]
\begin{center}
\resizebox{\linewidth}{!}{%
\begin{tabular}{rp{0.7cm}p{0.7cm}lp{1.9cm}}
\textbf{Name} & \textbf{Took \newline office}  & \textbf{Left \newline office} & \textbf{Party} & \textbf{Notes / \newline Events} \\
Benjamin Contee & 1789 & 1791 & Anti-Administration &  \\
William Pinkney & 1791 & 1791 & Pro-Administration & resigned\\
John Francis Mercer & 1792 & 1793 & Anti-Administration & \\
Uriah Forrest & 1793 & 1794 & Pro-Administration & resigned\\
Benjamin Edwards & 1795 & 1795 & Pro-Administration & \\
 \multicolumn{2}{c}{\large{$\vdots$}} & & & \\
\end{tabular}}
\begin{quote}
\begin{small}
What party was William Pinkney and Uriah Forrest a part of?\\
Answer: Pro-Administration\end{small}
\end{quote}
\end{center}
\caption{\label{tbl.example1}Example TableQA over Wikipedia Table}
\end{table}

Both the Row and Column models of RCI are sequence-pair classifiers. The question is one sequence and the text sequence representation of the row or column is the second sequence.
We consider two approaches to the sequence-pair classification task in RCI: Interaction and Representation. Interaction models use the self attention of a transformer over the concatenated two sequences. This is the standard approach to sequence-pair classification tasks, \eg{} textual entailment~\cite{Devlin2018BERTPO} \cite{wang2018glue}, in transformer based systems. 


Representation models independently project each sequence of the sequence-pair to a vector, then compare those vectors.
Representation models are motivated by the need to improve efficiency for a practical system. Considering the column classifier, the interaction model requires running a transformer over each question plus column sequence. In contrast, the representation model can pre-process the collection of tables, producing a vector representation of each column for each table, independent of any query.  Then, at query time, the query is projected to a vector which is then combined with the vector for each column and classified with a single-layer network. 
On the WikiTableQuestions-Lookup dev set, we see the column model's time drop from 40 seconds to 0.8 seconds on a K80 GPU when ten queries are batch processed at once. 

Let a table with $m$ rows and $n$ columns be defined as a header, $H = [h_1, h_2, ..., h_{n}]$ and cell values $V = [v_{i,j}], 1 \leq i \leq m, 1 \leq j \leq n$.  A TableQA instance consists of a table, a question and a ground truth set of cell indices, $T \subseteq I \times J, I = {1,2,...,m}, J= {1,2,...,n}$. In principle, these ground truth cell positions could be annotated with the correct occurrences of the correct values.
However, this form of supervision may be too difficult to obtain. We use \textit{weak supervision}: the ground truth cell indices are found by matching the ground truth answer strings in the table.
To train the row and column classifier we find ground truth row and column indices:

\begin{align*}
T_r & = \{i \mid \exists j : (i,j) \in T\} \\
T_c & = \{j \mid \exists i : (i,j) \in T\}
\end{align*}

Although it is possible to na\"ively construct a sequence representation of columns and rows by simply space separating the contents of each row or column, better performance can be achieved by incorporating the table structure in the sequence representation.  We focus on tables with a single header for columns, but this method could also be applied to tables with a hierarchical header, by first flattening the header.

The row ($S^{r}_{i}$) and column ($S^{c}_{j}$) sequence representations are formatted as:

\vspace{-3 mm}
\begin{align*}
S^{r}_{i} & = \bigoplus_{j=1}^{n} \zeta_h(h_j) \oplus\  \zeta_v(v_{i,j}) \\
S^{c}_{j} & = \zeta_h(h_j) \oplus \bigoplus_{i=1}^{m} \zeta_v(v_{i,j})
\end{align*}
\vspace{-4 mm}

Where $\oplus$ indicates concatenation and the functions $\zeta_h$ and $\zeta_v$ delimit the header and cell value contents. For $\zeta_h$ we append a colon token (`:') to the header string, and for $\zeta_v$ we append a pipe token (`$|$') to the cell value string. The particular tokens used in the delimiting functions are not important. Any distinctive tokens can serve since the transformer will learn an appropriate embedding to represent their role as header and cell value delimiters.

Considering again the example in Table~\ref{tbl.example1}, 
the first row would be represented as:
\begin{small}
\begin{quote}
Name : Benjamin Contee $|$ Took office : 1789 $|$ Left office : 1791 $|$ Party : Anti-Administration $|$ Notes / Events : $|$
\end{quote}
\end{small}

While the second column would have a sequence representation of:
\begin{small}
\begin{quote}
Took office : 1789 $|$ 1791 $|$ 1792 $|$ 1793 $|$ 1795 $|$
\end{quote}
\end{small}

Both the interaction and the representation models use the sequence representation described above.  In the case of the interaction model this sequence is then appended to the question with standard $[CLS]$ and $[SEP]$ tokens to delimit the two sequences. This sequence pair is then input to a transformer encoder, ALBERT. The final hidden state for the $[CLS]$ token is used in a linear layer followed by a softmax to classify the column as either containing the answer or not. 

\begin{figure}[thb]
   \centering
   \includegraphics[width=0.8\linewidth]{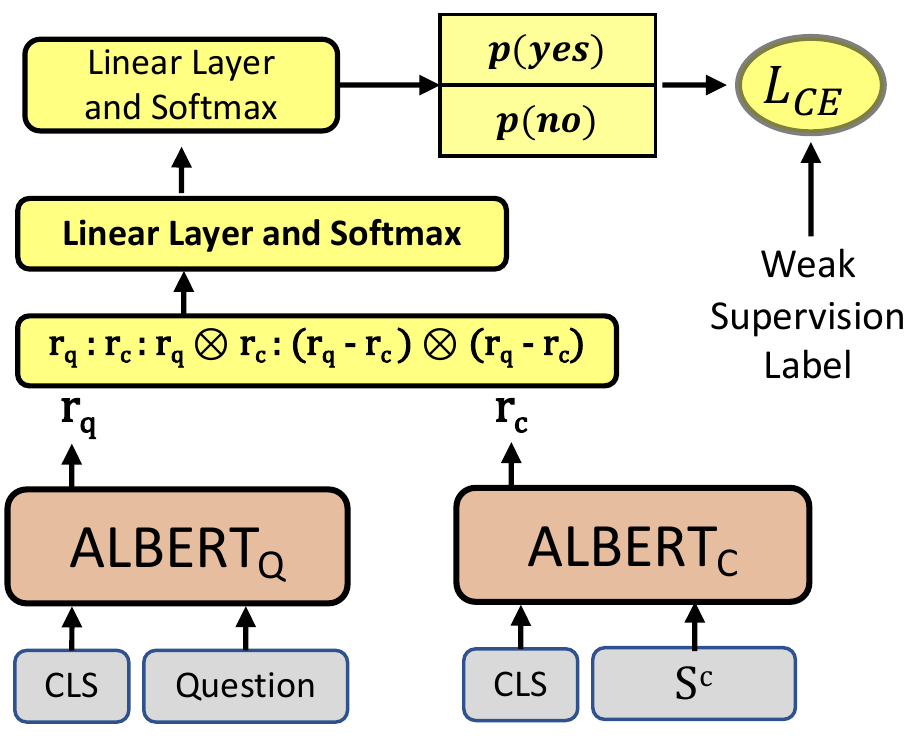}
   \caption{RCI Representation Model}
   \label{fig.RCIrepresentation}
\end{figure}

In the representation model shown in Figure~\ref{fig.RCIrepresentation} the representations of the question ($r_q$) and the $j$th column sequence ($r_{c}$) are first computed independently. The representations are taken from the vector that the transformer model produces for the $[CLS]$ input token. These vectors are then concatenated (indicated as $:$) with their element-wise product (indicated as $\otimes$) and the element-wise square of their differences. The probability that this column is the target for the question is then given by a softmax over a linear layer.

\begin{align*}
\mathbf{r_{\delta}} & = \mathbf{r_{q}} - \mathbf{r_{c}}\\
\mathbf{v_{qc}} & = \mathbf{r_{q}} : \mathbf{r_{c}} : \mathbf{r_{q}} \otimes \mathbf{r_{c}} : \mathbf{r_{\delta}} \otimes \mathbf{r_{\delta}} \\
p(j \in T_c) & = softmax(\mathbf{W} \mathbf{v_{qc}} + \mathbf{b})_0
\end{align*}


\noindent\textbf{Extension to aggregation questions:} Although our focus is on lookup questions, the RCI model can be extended to aggregation questions with the addition of a question classifier. Another transformer is trained to classify the sequence-pair of the question and the table header into one of six categories: lookup, max, min, count, sum and average. The table header is relevant because a question such as ``How many wins do the Cubs have?'' can be lookup, count or sum depending on the structure of the table.

Taking a threshold on the cell level confidences of the RCI model and aggregating by the predicted question type produces the final answer, either a list of cells for lookup questions or a single number for aggregation questions.

This approach requires \textit{full supervision}, we must know the cells to be aggregated to train the RCI row and column classifiers as well as the type of aggregation to train the question classifier. This type of supervision is available in the WikiSQL dataset, but not in WikiTableQuestions. 
\section{Evaluation}
\label{sec:eval}

\newcommand{\best}[1]{\textbf{#1}}

To evaluate these three approaches, we adapt three standard TableQA datasets: WikiSQL~\cite{zhongSeq2SQL2017}, WikiTableQuestions~\cite{pasupat2015compositional} and TabMCQ~\cite{jauhar2016tabmcq}.
WikiSQL and WikiTableQuestions include both lookup questions as well as aggregation questions. 
As mentioned in Section~\ref{sec:intro}, our primary focus in this paper is on \emph{lookup} questions that require selection and projection operations over tables (i.e., identifying the row and column of a table with very high precision for a given natural language question). We are releasing the processing and evaluation code for the datasets to support reproducibility\footnote{https://github.com/IBM/row-column-intersection}. Table~\ref{tbl.datasets} gives a summary of these datasets.

In WikiSQL, the ground truth SQL query is provided for each question, so questions involving an aggregation operation can be automatically excluded. The lookup questions are 72\% of the WikiSQL benchmark. 
WikiSQL has some questions ($<3\%$) with multiple answers. 
We treat these as a list of relevant items and use information retrieval metrics to measure the quality of a predicted ranked list of cells.

TabMCQ is a multiple-choice, lookup TableQA dataset over general science tables. We discard the multiple-choice setting and treat it as a standard open-ended QA task. 
However, some TabMCQ tables are very large. Of the 68 tables, 17 have more than 50 rows, with two tables containing over a thousand rows. We down-sample the rows that are not relevant for a given question, limiting the largest table size to 50 rows. Unlike the other two datasets, these tables are not Wikipedia tables and have an unusual format. A sample TabMCQ table is provided in the appendix.

WikiTableQuestions does not provide a definitive indication for what questions are lookup questions. To identify these questions we first filter questions with words indicating an aggregation, such as `average', `min', `max', etc. These questions were further filtered manually to get the WikiTableQuestions-Lookup set.

\begin{table}[t!]
\begin{center}
\begin{tabular}{rrrr}
\textbf{Dataset} & \textbf{Train}  & \textbf{Dev} & \textbf{Test} \\
\hline
WikiSQL &	40606 &	6017 &	11324 \\
TabMCQ &	5453 &	1819 &	1820 \\
WikiTableQuestions &	851 &	124 &	241
\end{tabular}
\end{center}
\caption{\label{tbl.datasets}Lookup TableQA Dataset Sizes}
\end{table}

In order to evaluate our proposed approaches on these datasets we built three different systems and also used three existing models: \semanticParsing, provided by~\cite{dasigi2019semparse}, \TaBERT~\cite{tabert} and \TAPAS~\cite{tapas}. \semanticParsing{} is a semantic parsing based model trained on WikiTablesQuestions~\cite{pasupat2015compositional} dataset (See Section~\ref{sec:relatedwork} for the details of this work). For building their model we used the code provided in~\cite{naacl2019url}.
For \TaBERT{} we trained the model for WikiSQL using the lookup subset, and for WikiTableQuestions we used the full training set and applied to the lookup subset.
For \TAPAS{} we used the trained BASE (reset) models\footnote{https://github.com/google-research/tapas} for WikiSQL and applied to the lookup subsets of the dev and test sets.


The \GAAMA{} and \GAAMAXXL{} models are based on Machine Reading Comprehension, using the \textit{base v2} and \textit{xxlarge v2} versions of ALBERT. Because this model returns a span rather than a cell prediction, we match each of the top-k span predictions to the closest cell, the cell with the lowest difference in its character offsets. In case multiple of the top-k predictions map to the same cell, these predictions are merged.

We also evaluate the two approaches to RCI: interaction (\RCIinteraction) and representation (\RCIrepresentation). Both models use the \textit{base v2} version of ALBERT.  
Using the \textit{xxlarge v2} ALBERT, we also train another RCI interaction model, \RCIXXL{}.  
For the representation model we found comparable performance on the column classifier but much lower performance on the row classifier.  Therefore the \RCIrepresentation{} model uses a representation based classifier for columns, while still using the interaction classifier for rows.  The \RCIinteraction{} model uses interaction classifiers for both rows and columns. Because WikiSQL is the largest dataset by far, for TabMCQ and WikiTableQuestions we first train models on WikiSQL, then fine tune on the target dataset. This gives small but significant gains for TabMCQ but is critical to good performance on WikiTableQuestions.

All models except \TAPAS{} produce a ranked list of top-k predictions. We evaluate these predictions using the metrics of Mean Reciprocal Rank (MRR) and Hit@1.
Mean Reciprocal Rank is computed by finding the rank of the first correct cell prediction for each question and averaging its reciprocal. If a correct cell is not present in the top-k predictions, it is considered to have an infinite rank.  Hit@1 simply measures the fraction of questions that are correctly answered by the first cell prediction.

\subsection{Results}

Table~\ref{tbl.mainResults} shows the results on the lookup versions of WikiSQL, TabMCQ, and WikiTableQuestions. Both the interaction and the representation models of RCI outperform all other methods on WikiSQL, TabMCQ, and WikiTableQuestions.
Using the representation model for the column classifier reduces performance by less than two percent on WikiSQL, and less than three percent on TabMCQ, but up to seven percent on WikiTableQuestions.

On two of the three datasets both \RCIinteraction{} and the more efficient \RCIrepresentation{} outperform \GAAMAXXL{} with far fewer parameters and computational cost.  Similarly, RCI with ALBERT-base outperforms even the large version of \TAPAS{} trained on WikiSQL, getting 94.6\% Hit@1 compared to the 89.43\% Hit@1 of \TAPAS$_{large}$.

\begin{table}[th!]
\small
\begin{center}
\resizebox{\linewidth}{!}{%
\begin{tabular}{rccccc}
 & \multicolumn{2}{c}{\textbf{Dev}} && \multicolumn{2}{c}{\textbf{Test}} \\ \cline{2-3}\cline{5-6}
\textbf{Model} & \textbf{MRR} & \textbf{Hit@1} && \textbf{MRR} & \textbf{Hit@1} \\ 
\hline \multicolumn{6}{c}{\textbf{WikiSQL-Lookup}} \\ \hline
\semanticParsing   & 0.752 & 67.11\% && 0.769 & 69.45\% \\
\GAAMA             & 0.766 & 66.91\% && 0.764 & 66.52\% \\
\TaBERT            & 0.759 & 70.78\% && 0.761 & 71.16\% \\
\TAPAS             & NA    & 91.32\% && NA    & 89.02\% \\
\RCIinteraction    & \best{0.963}	& \best{94.48\%} && \best{0.962} & \best{94.60\%} \\
\RCIrepresentation & 0.950 & 92.55\% && 0.948 & 92.72\% \\
\hline
\GAAMAXXL          & 0.893 & 84.89\% && 0.896 & 85.33\% \\
\TAPAS$_{large}$   & NA    & 92.02\% && NA    & 89.43\% \\
\RCIXXL            & \best{0.986} & \best{97.89\%} && \best{0.987} & \best{97.99\%}	\\	
\hline \multicolumn{6}{c}{\textbf{TabMCQ-Lookup}} \\ \hline
\semanticParsing   & 0.375 & 19.62\% && 0.301 &	16.86\% \\
\GAAMA         & 0.690 & 60.03\% && 0.679 & 59.29\% \\
\RCIinteraction    & \best{0.746} & \best{67.01\%} && \best{0.742} & \best{66.26\%} \\
\RCIrepresentation & 0.727 & 64.16\% && 0.725 & 63.74\% \\
\hline
\GAAMAXXL          & 0.708 & 63.00\% && 0.705 & 62.64\% \\
\RCIXXL            & \best{0.758} & \best{69.10\%} && \best{0.752} & \best{68.35\%} \\
\hline \multicolumn{6}{c}{\textbf{WikiTableQuestions-Lookup}} \\ \hline
\semanticParsing   & 0.663 & 58.87\% && 0.644 & 52.69\% \\
\GAAMA             & 0.681 & 58.87\% && 0.601 & 46.47\%\\
\TaBERT            & 0.686 & 61.29\% && 0.646 & 56.02\%\\
\RCIinteraction    & \best{0.734} & \best{66.94\%} && \best{0.708} & \best{61.83\%} \\
\RCIrepresentation & 0.708 & 62.90\% && 0.656 & 54.77\% \\
\hline
\GAAMAXXL          & 0.783 & 69.35\% && 0.732 & 64.73\% \\
\RCIXXL            & \best{0.796} & \best{72.58\%} && \best{0.794} & \best{72.20\%}	
\end{tabular}}
\end{center}
\caption{\label{tbl.mainResults} Results on TableQA Lookup Datasets}
\end{table}

\begin{table}[th!]
\begin{center}
\begin{tabular}{rcc}
\textbf{Model} & \textbf{Dev} & \textbf{Test} \\ \hline
\citet{wang2019learning} & 79.4\% & 79.3\% \\
\citet{min2019discrete} & 84.4\% & 83.9\% \\
\TAPAS$_{large}$  & 88.0\%    & 86.4\% \\
\TaBERT  & 70.53\%    & 70.94\% \\
\RCIXXL           & \best{89.7\%} & \best{89.8\%} \\
\end{tabular}
\end{center}
\caption{\label{tbl.tapasComparison} WikiSQL (including aggregation) accuracy}
\end{table}


We also compare the performance of the RCI model adapted to aggregation questions to the state-of-the-art \TAPAS{} reported results on WikiSQL. We use the evaluation script provided by \TAPAS{} to produce exactly comparable accuracy numbers for the full WikiSQL dataset. Table \ref{tbl.tapasComparison} shows the RCI model gains over three percent, even without table specific pre-training. It also outperforms \TaBERT{} model by a large margin of 18.86\%.


In Section~\ref{sec:model_rci} we described the method to transform a table into sequence representations of the rows and columns. We do an ablation study on the two larger datasets to understand the impact of incorporating table structure into the sequence representation relative to simply space separating the cell contents. Table~\ref{tbl.ablateFormatting} shows that we make moderate but significant and consistent gains with this approach, over two percent in Hit@1. 

\begin{table}[th!]
\begin{center}
\resizebox{\linewidth}{!}{%
\begin{tabular}{rccccc}
  & \multicolumn{2}{c}{\textbf{WikiSQL}} && \multicolumn{2}{c}{\textbf{TabMCQ}} \\ \cline{2-3}\cline{5-6}
\textbf{Model} & \textbf{MRR} & \textbf{Hit@1} && \textbf{MRR} & \textbf{Hit@1} \\ \hline
\RCIinteraction & \best{0.963} & \best{94.48\%} && \best{0.746} & \best{67.01\%} \\
-formatting &  0.947 &	92.26\% && 0.733 & 64.82\%
\end{tabular}}
\end{center}
\caption{\label{tbl.ablateFormatting} Results on Dev Sets, with formatting ablated}
\end{table}

We also decompose the performance of the tested systems in terms of row and column accuracy. The top predicted cell, if wrong, could have the wrong row, the wrong column, or both. Table~\ref{tbl.rowColumnResults} shows that predicting the correct column is generally easier than predicting the correct row. An interesting exception occurs with \GAAMA{} on the WikiSQL benchmark: the row prediction is more accurate than the column prediction. For the \GAAMA{} system, the table is a sequence of column headers, followed by a sequence of rows. Since the table is serialized in row-major order, all of the relevant information for a row is present locally, while the information for columns is distributed through the table sequence representation.


\begin{table}[th!]
\begin{center}
\resizebox{\linewidth}{!}{%
\begin{tabular}{rcccccc}
  & \multicolumn{2}{c}{\textbf{WikiSQL}} & \multicolumn{2}{c}{\textbf{TabMCQ}} & \multicolumn{2}{c}{\textbf{WTQ}} \\ 
\textbf{Model} & \textbf{Row} & \textbf{Col} & \textbf{Row} & \textbf{Col}  & \textbf{Row} & \textbf{Col} \\ \hline
\semanticParsing   & 83.1 & 82.1 & 70.1 & 41.5 & 62.2 & 82.2 \\
\GAAMA         & 85.2 & 73.8 & 64.6 & 90.5 & 56.4 & 78.8 \\
\RCIinteraction    & \best{96.7} & \best{98.0} & \best{73.6} & \best{92.2} & \best{64.3} & \best{92.1} \\
\RCIrepresentation & \best{96.7} & 96.0 & \best{73.6} & 89.0 & \best{64.3} & 85.1
\end{tabular}}
\end{center}
\caption{\label{tbl.rowColumnResults} Row/Column Accuracy Results on Test Sets}
\end{table}

The \RCIinteraction{} model is the best at both tasks, with \RCIrepresentation{} having the same performance at the row level task, since it uses the same model for rows.
The TabMCQ column level performance of \GAAMA{} is within two percent of \RCIinteraction{}, which may be surprising, especially considering its performance on WikiSQL. TabMCQ tables are constructed in an unusual way that permits high column prediction performance for an MRC system. The rows in TabMCQ have the structure of sentences, which is helpful for a system trained on the SQuAD and NQ reading comprehension tasks (Refer to the appendix for a sample TabMCQ table).

\subsection{Error Analysis}

To better understand the advantages and disadvantages of the Row-Column Intersection approach, we examine the 20 cases in the dev set of WikiTableQuestions-Lookup where \RCIinteraction{} does not provide the correct answer in first position but \GAAMAXXL{} does. We find nine cases where we could identify nothing that in principle prevents the \RCIinteraction{} model from answering correctly. We find seven cases where multiple rows need to be considered together, while the RCI models always consider rows independently.  WikiTableQuestions includes some questions like Table~\ref{tbl.example2}. Although the answer to this question is a cell in the table, it requires something like aggregation to answer. All rows for a given year must be checked to see if there is a `1st' in the Place column. This violates a key assumption of RCI: that rows may be examined independently. The final four cases also violate the assumptions of RCI. In two cases the answer is in the header of the table, while RCI assumes that it will be a cell. In one case the table extraction failed, and in the final case the question asks about the string length of one of the columns where the answer (8) happens to be in the table.

\vspace{4 mm}

\begin{table}[th!]
\begin{center}
\begin{small}
\begin{tabular}{rrrl}
\textbf{Season} & ... & \textbf{Discipline} & \textbf{Place} \\
\hline
2012 &&  Downhill     & 2nd \\
2012 &&  Downhill     & 3rd \\
\multicolumn{2}{c}{...} & & \\
2013 &&  Downhill     & 3rd \\
2013 &&  Super-G      & 1st \\
 \multicolumn{2}{c}{...} & & \\
2014 &&  Super-G      & 2nd \\
2014 &&  Super-G      & 1st \\
 \multicolumn{2}{c}{...} & & \\
\end{tabular}
\begin{quote}
In which year did Tina Weirather not earn 1st place?\\
Answer: 2012
\end{quote}
\end{small}
\end{center}
\caption{\label{tbl.example2}Example Multiple-Row Question}
\end{table}

We also examine the cases where \GAAMAXXL{} does not find the correct answer in first position but \RCIinteraction{} does. The most frequent error, occurring in eight of the seventeen cases, is a `near-miss'. Either \GAAMAXXL{} chooses a value from the wrong column in the right row or a value from the row before or after. This is illustrated in Table~\ref{tbl.gaama_error}, where \GAAMAXXL{} selects a value near the desired date that is easily confused with a location. In other cases a location from the previous or next row, which are adjacent in the input passage, can be selected instead.

\begin{table}[th!]
\begin{center}
\begin{small}
\resizebox{\linewidth}{!}{%
\begin{tabular}{rrl}
\textbf{Date} & \textbf{Opponent} & \textbf{Venue}  \\
\hline
\multicolumn{2}{c}{...} &  \\        
27 Aug 2005    & Wigan Athletic       & JJB Stadium        \\       
10 Sept 2005 & Chelsea              & Stamford Bridge    \\       
17 Sept 2005 & West Bromwich Albion & Stadium of Light   \\       
25 Sept 2005 & Middlesbrough        & Riverside Stadium  \\       
1 Oct 2005    & West Ham United      & Stadium of Light   \\       
\multicolumn{2}{c}{...} & 
\end{tabular}}
\begin{quote}
Where was the match on 17 September 2005 played?\\
Answer: Stadium of Light \\
\GAAMAXXL{} Answer: West Bromwich Albion
\end{quote}
\end{small}
\end{center}
\caption{\label{tbl.gaama_error} Example of Near-Miss by MRC}
\end{table}

We also conduct an error analysis of \RCIXXL{} on the first 50 aggregation questions it misses on the dev set of WikiSQL.  The largest category, with 24 cases, is correct answers by \RCIXXL{} counted wrong by mistakes in the ground truth. Usually (23) the ground truth indicates that there should be COUNT aggregation when no aggregation is correct. For example, ``What is the rank of manager Rob Mcdonald?'' where \textit{Rank} is one of the table columns is mistakenly indicated as a COUNT aggregation question.

The second largest category (9) occurs when the cells are ranked correctly, and the correct aggregation is predicted, but the threshold for choosing the cells to aggregate is too low (1) or too high (8).

Another common error (7) occurs when \RCIXXL{} predicts a lookup question with the answer in a similar numeric column when aggregation is required.  For example, the question
``How many votes were taken when the outcome was "6th voted out day 12"?'' is asked for a table with a \textit{Votes} column. \RCIXXL{} predicts it as a lookup question with the answer (``2-2-1 3-0'') from this column, while the ground truth is a COUNT aggregation.

The final significant category (7) is cases of questions that are unanswerable.  This can occur because the table does not contain an answer or because the answer cannot be computed from a SQL query, such as when the answer is a sub-string of a cell.

The final three error cases are: a wrong column is selected (the episode number in series rather than the episode number in season); the question ``What is the result when the 3rd throw is not 8?'' is interpreted as ``What is the result when the 3rd throw is \textit{something other than} 8?'' rather than the ground truth ``What is the result when the 3rd throw is \textit{literally `not 8'}?''; and  non-Latin characters must be matched to select the correct row.
\section{Conclusion}
\label{sec:conclusion}
 
In this paper we propose two novel techniques, RCI interaction and RCI representation, to tackle the problem of locating answers over tables for given natural language questions. These transformer based models are fine-tuned on ground truth tables to predict the probability of containing the answer to a question in the rows and columns of tables independently. These probabilities are either used to answer questions directly or highlight the relevant regions of tables as a heatmap, helping users to easily locate the answers over tables.

Our experiments prove that the RCI model outperforms the state-of-the-art transformer based approaches pre-trained on very large table corpora (\TAPAS~\cite{tapas} and \TaBERT~\cite{tabert}), achieving $\sim$3.4\% and $\sim$18.86\% additional precision improvement on the standard WikiSQL benchmark including both Lookup and Aggregation questions. 
The representation model, on the other hand, enables pre-processing the tables and producing the embeddings to store and further use during online query processing, providing significant efficiency advantages without compromising much on the accuracy of finding cell values in tables. As for the future work, we plan to explore the exploitation of domain-specific taxonomies and embeddings generated for domain-specific corpora to tackle the problem of answering natural language questions over tables in domains such as finance, aviation and health care.

\bibliography{main}
\bibliographystyle{acl_natbib}

\end{document}